%% file: neurips_2026.tex
\documentclass{article}

\usepackage[main, final]{neurips_2026}

\usepackage[utf8]{inputenc} 
\usepackage[T1]{fontenc}    
\usepackage{hyperref}       
\usepackage{url}            
\usepackage{booktabs}       
\usepackage{amsfonts}       
\usepackage{nicefrac}       
\usepackage{microtype}      
\usepackage{xcolor}         
\usepackage[table]{xcolor}
\usepackage{amsmath}    
\usepackage{amssymb}   
\usepackage{bm}   
\usepackage[most]{tcolorbox}
\usepackage{subcaption}
\usepackage{amsthm}
\usepackage{amsmath}
\usepackage{amssymb}
\usepackage{wrapfig}
\usepackage{tabularx}
\usepackage{array}       
\usepackage{makecell}    
\usepackage{color}
\usepackage{float}
\usepackage{multirow}
\usepackage{pifont}
\definecolor{searchblue}{RGB}{0, 180, 220}    
\definecolor{infoorange}{RGB}{200, 120, 50}   
\definecolor{thinkblue}{RGB}{50, 80, 160}     
\definecolor{answerred}{RGB}{180, 40, 60}     

\newcommand{\tokensearch}[1]{\textcolor{searchblue}{\texttt{<#1>}}}
\newcommand{\tokeninfo}[1]{\textcolor{infoorange}{\texttt{<#1>}}}
\newcommand{\tokenthink}[1]{\textcolor{thinkblue}{\texttt{<#1>}}}
\newcommand{\tokenanswer}[1]{\textcolor{answerred}{\texttt{<#1>}}}

\title{PiCA: Pivot-Based Credit Assignment for Search Agentic Reinforcement Learning}

\author{%
   Dongyi Liu$^1$\thanks{Equal contribution.}\ ,\,
  Yifan Niu$^1$$^*$,\, 
  Qinwen Wang$^1$,\, 
  Han Xiao$^1$,\, 
  Jia Li$^{1,2}$\thanks{Correspondence to: Jia Li (\texttt{jialee@ust.hk}).}  \\ \\
  $^1$ The Hong Kong University of Science and Technology (Guangzhou) \\
  $^2$ The Hong Kong University of Science and Technology \\
}

\begin{document}

\maketitle

\begin{abstract}
Large Language Model (LLM)-based search agents trained with reinforcement learning (RL) have significantly improved the performance of knowledge-intensive tasks. However, existing methods encounter critical challenges in long-horizon credit assignment: (i) Reward Sparsity, where models receive only outcome feedback without step-level guidance to differentiate action quality; (ii) Isolated Credit, where credit is assigned to steps independently, failing to capture sequential dependencies; and (iii) Distributional Shift, where rewards are estimated on templates that deviate from the model’s natural generative distribution. To address these issues, we propose \textbf{Pi}vot-Based \textbf{C}redit \textbf{A}ssignment (PiCA), a novel step reward mechanism that reformulates the search trajectory as a sequential process of cumulative search progress. 
Unlike prior isolated step rewards, PiCA defines process rewards as success probabilities dependent on the historical context based on Potential-Based Reward Shaping (PBRS). This approach identifies pivot steps, which comprise target golden sub-queries and sub-answers derived from historical trajectories, as information peaks that significantly boost the likelihood of a correct final answer. By anchoring these step rewards to the final task objective, PiCA provides dense, pivot-aware and trajectory-dependent guidance while maintaining distributional consistency. Extensive experiments show that PiCA outperforms existing strong baselines across seven knowledge-intensive QA benchmarks, achieving 15.2\% and 4.4\% improvements for 3B and 7B models. The consistent performance gains across various models show PiCA’s robust generalization. 
The code is available at {\url{https://github.com/novdream/PiCA}}.

\end{abstract}

\section{Introduction}

Large Language Model~(LLM)-based search agents~\cite{searchr1,stepsearch,wang2026informationgainbasedpolicyoptimization} have recently redefined the paradigm for addressing long-horizon, knowledge-intensive tasks, such as multi-hop question answering~\cite{chen2025researchlearningreasonsearch,shi2025searchrefinethinkfacilitating} and open-domain information seeking~\cite{zheng2025deepresearcherscalingdeepresearch,zhao2026trainingmultiturnsearchagent}. For example, popular search agents, such as WebDancer~\cite{wu2025webdancerautonomousinformationseeking}, WebLeaper~\cite{tao2025webleaperempoweringefficiencyefficacy} and MiroThinker~\cite{miromindteam2026mirothinkerpushingperformanceboundaries}, can autonomously refine queries, summarize retrieved information from external environments through search tools~(\emph{e.g.}, online API, local corpus).

A primary bottleneck in these long-horizon tasks lies in incorrect credit assignment~\cite{lin2025comprehensivesurveyreinforcementlearningbased,tan2026hindsightcreditassignmentlonghorizon,zhang2026reasoningagenticcreditassignment}, specifically the misattribution of rewards to less important steps. Precise credit assignment is thus crucial for enabling search agents to decide when to search, how to formulate or refine queries, and how to incorporate retrieved evidence into multi-step reasoning.

Recently, Reinforcement Learning (RL)~\cite{RL} has emerged as a promising paradigm for developing adaptive and autonomous search agents with verifiable rewards. However, existing methods face three primary limitations:
\textbf{(1) Sparse Rewards}. Early efforts~\cite{searchr1,sun2025zerosearch} primarily rely on outcome-only supervision (\emph{i.e.}, answer accuracy). These approaches often lead to biased reward estimation, where high answer rewards are mistakenly given to wrong intermediate steps~(\emph{e.g.}, redundant query turns, wrong sub-answer turns).
\textbf{(2) Isolated Credit Assignment}. While new methods~\cite{stepsearch,xie2026tipsturnlevelinformationpotentialreward} attempt to incorporate fine-grained step rewards, they often suffer from isolated credit assignment where the step reward is estimated based solely on the local quality of the current turn. This neglects the inherent nature of knowledge-intensive tasks as a Markov Decision Process (MDP)~\cite{lin2025comprehensivesurveyreinforcementlearningbased}.
\textbf{(3) Distributional Shift}. Recent methods~\cite{wang2026informationgainbasedpolicyoptimization,xie2026tipsturnlevelinformationpotentialreward} concatenate ground-truth data with intermediate steps to estimate answer probability to give dense step rewards. However, this concatenation triggers a distributional shift since such sequences are absent from the model's natural generation, resulting in biased rewards.

To address these limitations, we propose \textbf{Pi}vot-Based \textbf{C}redit \textbf{A}ssignment (PiCA), which reformulates the search trajectory as a sequential process of cumulative search progress to reach the final correct goal. Unlike prior works that reward steps in isolation, we define the process reward as a success probability dependent on the historical trajectory based on potential-based reward shaping (PBRS)~\cite{Wiewiora_2003,ng_harada_russell_1999}. In our formulation, PiCA follows a core intuition:
\begin{tcolorbox}[colback=white,colframe=gray!50!black,title={Main Intuition}] %
    \centering
    \renewcommand{\arraystretch}{1}
    \begin{tabular}{p{1.0\linewidth}}
        The probability of reaching the correct answer in multi-hop tasks increases with the cumulative acquisition of \textbf{pivot steps}~(\emph{i.e.}, deriving target golden sub-queries and sub-answers based on historical trajectory)
    \end{tabular}
\end{tcolorbox}
Under this formulation, each pivot step represents a distinct peak in search progress, directly increasing the  success probability of the final goal. Conversely, recognizing that non-pivot steps do not necessarily compromise the final goal, we explicitly link each process reward to the final outcome via conditional probability to better align with the final task objective. By augmenting PPO with our turn-level rewards, extensive experiments across seven in-domain and out-of-domain QA benchmarks confirm the efficacy of our method in retrieving relevant information and  synthesizing evidence into correct, verifiable answers in knowledge-intensive tasks. Our main contributions are two folds: (1) We propose \textbf{PiCA}, a novel credit assignment framework where step rewards are dependent on the entire historical trajectory and can reflect which step has effective information. (2) PiCA outperforms competitive baselines by 15.2\% (3B) and 2.2\%  (7B) average on seven multi-hop QA benchmarks.

\section{Related Work}
\subsection{Retrieval-Augmented Generation}
Retrieval-Augmented Generation (RAG) enhances generative models by incorporating external knowledge, overcoming the limitations of static parameters \cite{lewis2020retrieval}. Current research can be categorized into three stages based on the depth of integration between retrieval and reasoning. First, prompt-driven augmentation methods leverage prompt engineering to guide models through query decomposition and multi-turn retrieval. Representative works like REPLUG \cite{shi2024replug} align retrievers with black-box models, while FLARE \cite{jiang2023active} introduces active retrieval based on generation confidence. However, these methods are often susceptible to interference from redundant information in long contexts \cite{liu2024lost}. Second, fine-tuning and reflection-based architectures employ supervised fine-tuning (SFT) to empower models with self-critique capabilities. Self-RAG \cite{asai2023self} utilizes reflection tokens for iterative optimization, CRAG \cite{yan2024corrective} introduces corrective retrieval mechanisms, and RetroLLM \cite{li2025retrollm} focuses on fine-grained evidence extraction to improve information utilization. Third, inference-time scaling and search-based methods, inspired by reasoning models like o1, focus on increasing computational investment during inference. RAG-star \cite{jiang2025rag} and AirRAG \cite{feng2025airrag} utilize Monte Carlo Tree Search (MCTS) for path exploration, while Search-o1 \cite{li2025search} models retrieval as an agentic behavior, significantly enhancing planning capabilities in complex tasks. Despite these advances, RAG still faces core bottlenecks: cascading errors where early missteps propagate through the reasoning chain, noise interference in long contexts affecting precise evidence extraction, and high computational costs or latency associated with search-augmented methods. 

\subsection{Reinforcement Learning for Agentic Search}
Reinforcement learning (RL) has become a key component in post-training large language models for reasoning and decision-making. Outcome-supervised paradigms, such as RLHF and PPO-style methods (e.g., GRPO), have demonstrated strong performance on verifiable tasks including mathematics and code generation~\cite{ouyang2022rlhf, schulman2017ppo, shao2024grpo}. However, these approaches primarily rely on final outcome signals, which leads to a fundamental credit assignment challenge in long-horizon reasoning tasks: the model lacks explicit feedback on how intermediate reasoning steps contribute to the final result~\cite{arjona2019rtr}.To address this limitation, prior work has explored optimizing search and reasoning behaviors within an RL framework. For instance, Search-R1 formulates search as a reinforcement learning environment to learn query generation strategies~\cite{searchr1}, while StepSearch introduces step-wise rewards based on search progress to guide retrieval behavior~\cite{stepsearch}. TIPS further constructs dense rewards based on improvements in answer likelihood to stabilize long-horizon training~\cite{tips}. In addition, Zerosearch simulates search environments and controls retrieval quality, enabling efficient training without relying on real-world search APIs~\cite{sun2025zerosearch}.Despite these advances, existing methods still lack explicit modeling of step dependencies and the evolution process in multi-step reasoning, making it difficult to capture reasoning deviations and step-level contributions, thereby limiting fine-grained credit assignment in complex multi-hop scenarios.

\section{Preliminary}
In this section, we first establish the search agent task formulation in Section~\ref{sec:task_formulation}, followed by a detailed description of the search agentic RL pipeline in Section~\ref{sec:agentic_rl}.

\subsection{Task Formulation}\label{sec:task_formulation}
Given a question $q$, a language model $\pi_\theta$ generates a trajectory $y = (\tau_1, \tau_2, \dots, \tau_T)$, where $T$ denotes the total number of interaction turns. Each turn $\tau_t$ is defined as a composite semantic block consisting of three functional components: a reasoning step~\tokenthink{think}, a tool invocation~\tokensearch{search}, and an environment observation ~\tokeninfo{information}. The entire trajectory is assigned a binary label $l \in \{0, 1\}$, which serves as the ground-truth signal extracted from the last turn in ~\tokenanswer{answer} tag. 

We formulate the agentic search process as a MDP denoted by the tuple $\mathcal{M}=(\mathcal{S},\mathcal{A},\mathcal{P},r,\gamma)$. 
At each turn $t$, $s_t=(q,\tau_{\leq t-1})$ contains the question $q$ and the sequence of previously generated interaction turns $\tau_{\leq t-1}$.
Based on this state, $\pi_\theta$ generates actions $a_t$ following the format 
$a_t = ~\tokenthink{think} \tokensearch{search}$.
Upon executing the think and search action $a_t$, the external search engine tools $\mathcal{T}$ returns an observation $o_t$ with ~\tokeninfo{information} tag. The interaction turn is then defined by their concatenation, $\tau_t = a_t \oplus o_t$.
The state transition is deterministic, as the next state is uniquely determined by concatenating the previous sequence with the current output $a_t$. $r_t$ denotes the reward provided by the environment or generated by the reward model for action $a_t$.

For PiCA training, we denote the training dataset as $\mathcal{D} = \{(q_i, y_i, l_i, \mathbf{z}_i)\}$, where $q_i$ is the $i$-th question, $y_i$ is the corresponding response consisting of a multi-step reasoning trajectory, $l_i$ indicates whether the final answer is correct, and $\mathbf{z}_i$ is a sequence of binary labels where each $z_{i,j}$ signifies whether the $j$-th step in $y_i$ is a pivot step. Formally, we define $\mathcal{D}_p = \{s_{i,j} \in y_i \mid z_{i,j} = \text{true}\}$ as the set of all pivot steps identified across all trajectories in the dataset $\mathcal{D}$, while the detailed construction prompts and criteria for these pivot labels are described in Appendix~\ref{app:generation}.

\subsection{Proximal Policy Optimization with Search Engine}\label{sec:agentic_rl}
Compared to standard PPO~\cite{schulman2017proximalpolicyoptimizationalgorithms}, at each turn $t$, the search agent's trajectory includes external environmental observation $o_t$. To ensure gradients are only propagated through the model's policy, we employ a token-level mask $I(y_{t})$ to isolate model-generated actions $a_t$ from the observations $o_t$.

\begin{equation}\label{eq:ppo_search}
\begin{aligned}
&\mathcal{J}_{\text{PPO}}(\theta) =\mathbb{E}_{q \sim \mathcal{D}, y \sim \pi_{\theta_{\text{old}}}(\cdot|q)} \\
   &\Bigg\{ \frac{1}{\sum_{t=1}^{|y|} I(y_{t})} \sum_{t=1:I(y_{t})=1}^{|y|} \Bigg[ \frac{\pi_{\theta}(y_{t}|q,y_{<t})}{\pi_{\theta_{\text{old}}}(y_{t}|q,y_{<t})} A_{t}, 
 \text{clip}\Bigg( \frac{\pi_{\theta}(y_{t}|q,y_{<t})}{\pi_{\theta_{\text{old}}}(y_{t}|q,y_{<t})} , 1-\epsilon, 1+\epsilon \Bigg)A_{t} \Bigg] \Bigg\},
\end{aligned}
\end{equation}
where \( \pi_{\theta} \) and \( \pi_{\text{old}} \) represent the current and previous policy models, respectively. 
$I(y_t)=1$ if $y_t$ is a LLM generated token~(\emph{i.e.}, \tokenthink{think}, \tokensearch{search}) and $I(y_t)=0$ if $y_t$ is a retrieved token~(\emph{i.e.}, \tokeninfo{information}). The term \( \epsilon \) is a clipping-related hyperparameter introduced in PPO to stabilize training. 
Following Generalized Advantage Estimation (GAE), $A_t$ is estimated using $V_{\phi}$ and future rewards $\{ r_{\geq t} \}$ derived from both the final answer and the reward model (Section~\ref{sec:reward_model}).

\begin{figure}[t]
    \centering
     \includegraphics[width=1.0\linewidth, height=0.55\linewidth, keepaspectratio=false]{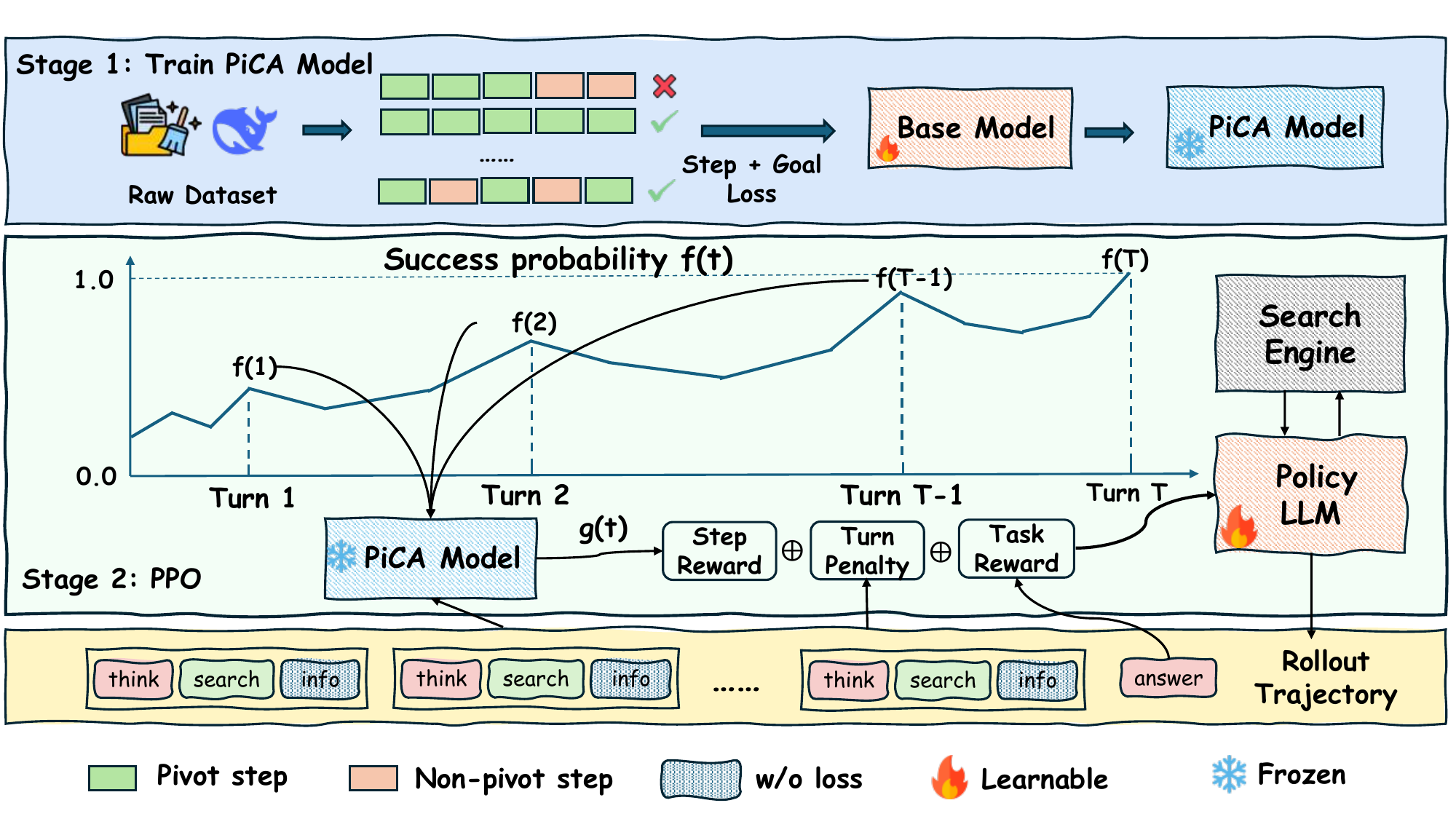}
    \caption{Overview of PiCA. Stage 1 is training a PiCA model on annotated pivot steps (Section 4.1). Stage 2 is policy optimization with PiCA (Section 4.2). The model generates trajectories with a frozen PiCA model assigning dense rewards based on relative success gain $g(t)$ derived from the success probability $f(t)$. The total reward integrates these step-wise gains, efficiency penalties, and final task outcomes, while retrieved content is masked during training. }
    \vspace{-10pt}
    \label{fig:placeholder}
\end{figure}

\section{Methodology}
In this section, we detail the description of PiCA (Section~\ref{sec:reward_model}). Building on this, we detail the design of the reinforcement learning algorithm with PiCA (Section~\ref{sec:optimization}).

\subsection{Pivot-based Credit Assignment}\label{sec:reward_model}

To provide fine-grained supervision, we model the search trajectory as the evolution of the probability of reaching a correct answer based on historical states. In multi-step search, identifying an intermediate step as a fatal error is often ambiguous, since the model's self-correction mechanism~\cite{wang2024theoreticalunderstandingselfcorrectionincontext,kumar2024traininglanguagemodelsselfcorrect} allows it to recover from suboptimal states and eventually converge to the solution. However, the acquisition of pivot steps (\emph{i.e.}, deriving target golden sub-queries and sub-answers based on historical trajectory) provides a clear signal of progress. To this end, our PiCA model generates step rewards based on the increment in success probability rather than the likelihood of failure. We define the success probability $f(t)$ as:
\begin{equation}
f(t) = P(l=1 \mid  s_t, a_t)
\end{equation}
where $l=1$ indicates the final answer is correct. The initial probability $f(0) = P(l=1 \mid x)$ represents the prior difficulty of the problem given only the question. At the final step $T$, the state converges to a deterministic outcome: $f(T) = 1$ if the answer is correct, and $f(T) = 0$ otherwise.

\textbf{Relative Success Gain}. To measure the improvement in success likelihood elicited by the acquisition of new information, we calculate the relative success gain $g(t)$ based on the preceding steps in the trajectory. This metric quantifies the normalized change in success probability:
\begin{equation}
g(t) = \frac{f(t) - f(t-1)}{f(t-1)} = \frac{\Delta f(t)}{f(t-1)}  
\end{equation}
Here, $g(t) > 0$ implies the step is a productive advancement that raises the probability of success, while $g(t) < 0$ indicates the introduction of errors or logical confusion. Consequently, the success probability at any turn $t$ can be decomposed as a product of these relative gains:
\begin{gather}
f(t) = f(0) \cdot \frac{f(1)}{f(0)} \cdot \frac{f(2)}{f(1)} \cdots \frac{f(t)}{f(t-1)} = f(0) \prod_{k=1}^t \left( 1 + \frac{f(k) - f(k-1)}{f(k-1)} \right) \\
= f(0) \prod_{k=1}^t (1 + g(k)) \notag
\end{gather}

\textbf{Reward Shaping for Dense Signal}. We employ Potential-Based Reward Shaping (PBRS) to map these probabilities into a dense reward signal. We define the potential function $\Phi(s_t)$ as the logarithm of the  success probability:
\begin{equation}
    \Phi(s_t) \equiv \log f(t) 
\end{equation}

Following the PBRS framework, the shaped reward $r_t$ for the transition from $s_{t-1}$ to $s_t$ is given by:
\begin{equation}
    r_t \equiv R(s_{t-1}, a_{t-1}, s_t) + \gamma \Phi(s_t) - \Phi(s_{t-1})
\end{equation}
By setting the intermediate environmental reward $R = 0$ and the discount factor $\gamma = 1$, the process reward simplifies to the log-ratio of relative success gain:
\begin{equation}
\begin{aligned}
r_t &= \Phi(s_t) - \Phi(s_{t-1}) = \log f(t) - \log f(t-1) \\
&= \log \left( \frac{f(t)}{f(t-1)} \right) = \log \left( \frac{f(t-1) + \Delta f(t)}{f(t-1)} \right) = \log \left( 1 + \frac{\Delta f(t)}{f(t-1)} \right) \\
&= \log(1 + g(t))
\end{aligned}
\end{equation}

\begin{wrapfigure}{r}{0.54\textwidth}  
    \centering
    \vspace{-10pt}
    \includegraphics[width=\linewidth]{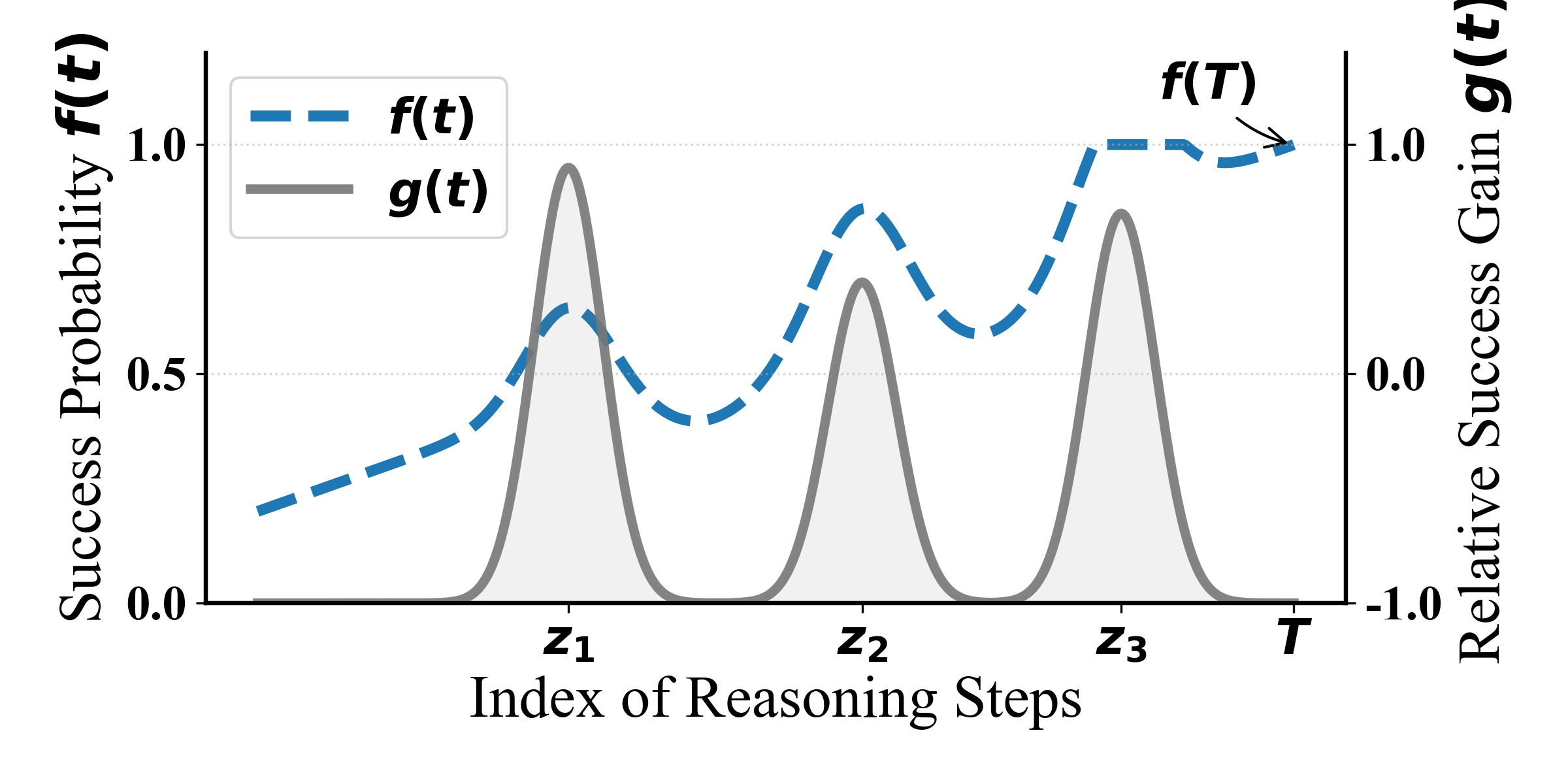}  
    \caption{PiCA Step Reward}  
    \label{fig:reward_steps}
\end{wrapfigure}

\textbf{Reward Model Training}. Based on the dataset $\mathcal{D}$ and pivot steps $\mathcal{D}_p$ and the success reward defined in Eq. (7), our training objective is designed to maximize the rewards of pivot steps while enabling the model to autonomously reward intermediate steps through final outcomes. The total loss consists of two components.

\textbf{Step-level Explicit Supervision}. For the pivot steps $t \in \mathcal{D}_P$ within each trajectory $ \in \mathcal{D}$, we define the loss $\mathcal{L}_{\text{gold}}$ to ensure these actions provide positive search progress ($g_t > 0$). This explicitly encourages the model to recognize pivot steps as milestones that advance the state toward correctness:
\begin{equation}
    \mathcal{L}_{\text{gold}} = - \sum_{L \in \mathcal{D}} \sum_{t \in \mathcal{D}_p} \log (g_t)
\end{equation}

\textbf{Outcome-level Implicit Supervision}.
For the complete trajectory $y$, we utilize the outcome label $l \in \{0, 1\}$ to supervise the final success probability $f(T)$. This objective allows the model to autonomously reward non-pivot steps based on whether the overall search process converges to a correct solution. The outcome-level loss is formulated as:
\begin{equation}
    \mathcal{L}_{\text{final}} = \begin{cases} -\log f(T), & \text{if } l = 1 \\ -\log(1 - f(T)), & \text{if } l = 0 \end{cases}
\end{equation}
where $f(T) \to 1$ for successful trajectories and $f(T) \to 0$ for failures.
The overall loss to be minimized is as follows:
\begin{equation}
    \mathcal{L} = \frac{1}{|\mathcal{D}|} \sum^{{\mathcal{D}}} (\mathcal{L}_{\text{final}} + \lambda_g \cdot \mathcal{L}_{\text{gold}})
\end{equation}
where $\lambda_g$ is the weight coefficient to balance the two loss terms. 

\subsection{Policy Optimization}\label{sec:optimization}

We optimize the search policy $\pi_\theta$ using Proximal Policy Optimization (PPO) in Eq.~1. The objective is to maximize the expected return by calculating the advantage $\hat{A}_t$, which depends on the value function $V_\phi$ and the combined reward signal. In our method, the total reward $R$ for a search trajectory is decomposed into two components: the final outcome reward ($r_{out}$) and the step reward ($r_{step}$).

\textbf{Outcome Reward ($r_{out}$)}. To ensure the model adheres to structural constraints and reaches the correct solution, the outcome reward is applied only at the final token of the trajectory, which consists of a format reward and a F1 answer reward.

\textbf{PiCA (${r}_{step}$)}.
The step-wise reward is assigned to the last token of each search behavior (\emph{i.e.}, \tokensearch{search}) round to provide dense supervision. However, a naive summation of rewards may trigger reward hacking~\cite{skalse2025definingcharacterizingrewardhacking,wang2026rewardhackingeralarge}. To mitigate this, we introduce a step penalty for trajectories:
\begin{equation}
r_{step, t} =
\begin{cases}
PiCA(s_t, a_t), & \text{if } t < 3 \\
 PiCA(s_t, a_t) - \lambda \cdot \alpha^{(t-3)}, & \text{if } t \geq 3
\end{cases}
\end{equation}
where $\alpha \in [1, 1.5]$ is the exponential growth factor and $\lambda \in [0, 0.5]$ is the base penalty coefficient. This penalty ensures  the model is encouraged to perform necessary search steps.

\textbf{Policy Optimization}. For a trajectory $y = (\tau_1, \tau_2, \dots, \tau_T)$, we define the turn reward $R_t$ as \textbf{the last token}~(\emph{i.e.}, \tokensearch{search}, \tokenanswer{answer}) in turn $t$. Intermediate turns receive $r_{step, t}$, while the final turn $T$ incorporates outcome reward $r_{\text{out}}$ to signal the overall success or failure of the trajectory to the model. 
\begin{equation}
    R_t = \begin{cases} 
r_{step, t}, & \text{if } t < T \\ 
r_{step, T} + r_{\text{out}}, & \text{if } t = T 
\end{cases}
\end{equation}

The turn-level advantage is $A_t = R_t + \gamma V_\phi(s_{t+1})-V_\phi(s_t)$, where $V_{\phi}(s_t)$ is a baseline value predicted by critic model. To incorporate such long-horizon dependencies, we compute a discounted cumulative advantage to propagate outcome signals backward to earlier turns.
\begin{equation}
\tilde{A}_{t}=\sum_{l=0}^{T-t}(\gamma\lambda)^lA_{t+l}.
\end{equation}
where $\gamma \in (0,1]$ is the discount factor, and $\lambda \in [0,1]$ controls the propagation of future advantage signals. For any token $j$ generated during turn $t$, its assigned token-level advantage is $ \tilde{A}_{(j)} = \tilde{A}_{t} $. With the discounted advantages $ \tilde{A}_{(j)}$ defined above, we optimize the agent policy following the same structure as PPO but with a finer-grained credit assignment. Formally, let $r_j(\theta) = \frac{\pi_{\theta}(y_{j}|q,y_{<j})}{\pi_{\theta_{\text{old}}}(y_{j}|q,y_{<j})}$, the final objective is
\begin{equation}
\begin{aligned}
\mathcal{J}_{\text{PiCA}}(\theta) = \mathbb{E}_{q, y} \Bigg[ \frac{1}{\sum I(y_{j})} \sum_{j=1}^{|y|} I(y_j) \cdot \min \left( r_j(\theta) \tilde{A}_{(j)}, \text{clip}(r_j(\theta), 1-\epsilon, 1+\epsilon) \tilde{A}_{(j)} \right) \Bigg]
\end{aligned}
\end{equation}

\vspace{-10pt}
\section{Experiments}
\vspace{-5pt}

\subsection{Experimental Setup}\label{sec:experimental setup}
\textbf{Reward Model Training.}
We train Qwen2.5-3B-Instruct on new MuSiQue dataset, which contains approximately 60,000 trajectories, along with step-level process annotations. The reward model is trained via full-parameter fine-tuning. More training details are provided in Appendix~\ref{app:implementation}.

\textbf{Search Agent Training.}
Following a multi-turn question answering setup, we conduct experiments with two model scales, Qwen-2.5-3B-Instruct and Qwen-2.5-7B-Instruct~\cite{qwen2025technical}. For retrieval, we adopt the E5 encoder~\cite{wang2022e5} over the 2018 Wikipedia corpus, retrieving 3 documents at each interaction step. The models are trained on a combined dataset constructed from the NQ and HotpotQA training splits. We evaluate both in-domain and out-of-domain performance on seven QA benchmarks: NQ, TriviaQA, PopQA, 2WikiMultiHopQA, MuSiQue, HotpotQA, and Bamboogle~\cite{joshi2017triviaqa,kwiatkowski2019natural,mallen2023popqa,ho2020twowiki,trivedi2022musique,yang2018hotpotqa,press2023bamboogle}. Performance is measured using Exact Match (EM) and F1 scores~\cite{jin2025empiricalstudyreinforcementlearning}.

\textbf{Implementation Details.} We adopt a unified set of hyperparameters across all methods. We use PPO with GAE ($\lambda=1$, $\gamma=1$), a KL penalty coefficient $\beta=0.001$, and a clipping ratio $\epsilon=0.2$. The batch size is set to 256, with a maximum context length of 4096 tokens and up to 4 retrieval turns per query. Training is conducted for 190 steps, or until performance collapses, using 8 $\times$ A800 GPUs with FSDP and gradient checkpointing. Additional details on training hyperparameters and the search engine server configuration are provided in Appendix~\ref{app:implementation}.

\textbf{Baselines.}
We compare our method with a set of representative reinforcement learning approaches for search-augmented reasoning, including RAG, Search-o1, Search-R1, ZeroSearch, StepSearch, TIPS and MT-PPO. To ensure fair comparison, we follow prior work in adopting the same multi-turn question answering framework and retrieval configurations. All baselines are implemented following their original configurations, and we report the average performance for each model size.

\vspace{-5pt}
\subsection{Main Results}\label{sec:main results}
\vspace{-5pt}
\begin{table*}[!t]
\caption{Exact Match (EM) results on seven QA benchmarks. In-domain tasks include NQ and HotpotQA; out-of-domain tasks include TriviaQA, PopQA, 2WikiMultiHopQA, MuSiQue, and Bamboogle.}
\centering
\small
\setlength{\tabcolsep}{6pt}
\begin{tabular}{l | cc | ccccc | c}
\toprule
\multirow{2}{*}{Method}
& \multicolumn{2}{c|}{In-domain}
& \multicolumn{5}{c|}{Out-of-domain}
& \multirow{2}{*}{Avg} \\
\cmidrule(lr){2-3}
\cmidrule(lr){4-8}
& NQ & HotpotQA
& TriviaQA & PopQA & 2Wiki & MuSiQue & Bamboogle
& \\
\midrule

\multicolumn{9}{c}{\textbf{Qwen2.5-3B Instruct}} \\
\midrule
RAG         & 0.348    & 0.251 & 0.544    & 0.387    & 0.221 & 0.051 & 0.076 & 0.283 \\
Search-o1   & 0.238 & 0.240 & 0.472 & 0.262 & 0.207 & 0.045 & 0.316 & 0.254 \\
Search-R1   & 0.341 & 0.324 & 0.545 & 0.378 & 0.319 & 0.103 & 0.264 & 0.325 \\
ZeroSearch  & 0.414 & 0.267 & 0.574 & \textbf{0.448} & 0.239 & 0.088 & 0.193 & 0.318 \\
StepSearch  & --    & \underline{0.345} & --    & --    & \underline{0.320} & \textbf{0.174} & \underline{0.344} & -- \\
MT-PPO      & 0.397 & 0.255 & 0.562 & 0.405 & 0.214 & 0.062 & 0.080 & 0.282 \\
TIPS        & \textbf{0.435} & 0.314 & \underline{0.588} & \underline{0.428} & 0.293 & 0.087 & 0.208 & \underline{0.336} \\

\midrule
\textbf{Ours}
& \underline{0.426} & \textbf{0.400}
& \textbf{0.612} & 0.417 & \textbf{0.408} & \underline{0.160} & \textbf{0.347}
& \textbf{0.396} \\

\midrule
\multicolumn{9}{c}{\textbf{Qwen2.5-7B Instruct}} \\
\midrule
RAG         & 0.349    & 0.287 & 0.585    & 0.392    & 0.231 & 0.061 & 0.214 & 0.283 \\
Search-o1   & 0.151 & 0.193 & 0.443 & 0.131 & 0.181 & 0.053 & 0.302 & 0.208 \\
Search-R1   & 0.393 & 0.370 & 0.610 & 0.397 & \underline{0.401} & 0.146 & 0.368 & 0.385 \\
ZeroSearch  & \underline{0.436} & 0.325 & 0.618 & \textbf{0.515} & 0.309 & 0.120 & 0.267 & 0.370 \\
StepSearch  & --    & 0.386 & --    & --    & 0.366 & \textbf{0.226} & \underline{0.400} & -- \\
MT-PPO      & 0.424 & 0.265 & 0.551 & 0.416 & 0.228 & 0.069 & 0.112 & 0.295 \\
TIPS        & 0.434 & \underline{0.421} & \underline{0.640} & \underline{0.450} & \textbf{0.430} & 0.170 & 0.368 & \underline{0.417} \\
\midrule
\textbf{Ours}
& \textbf{0.460} & \textbf{0.424}
& \textbf{0.641} & 0.442 & \underline{0.401} & \underline{0.197} & \textbf{0.419}
& \textbf{0.426} \\

\bottomrule
\end{tabular}
\vspace{-10pt}
\label{tab:main_results}
\end{table*}

In Table~\ref{tab:main_results} and \ref{tab:main_results_f1}, we compare with other competitive prompt-based and RL-based baselines on seven standard benchmarks (\emph{i.e.}, in-domain, out-of-domain) to validate the effectiveness  of our method. The EM and F1 scores are reported as the average of three independent runs. Based on these results, we can draw the following key observations:

\textbf{(1) PiCA demonstrates excellent performance in knowledge-intensive tasks}. As shown in Tables \ref{tab:main_results} and \ref{tab:main_results_f1}, Ours consistently outperforms competitive baselines on in-domain benchmarks. On Qwen2.5-3B, our method achieves the highest EM (0.400) and F1 (0.514) on HotpotQA, significantly surpassing models like StepSearch and Search-R1. It also maintains superior accuracy on NQ with a peak F1 score of 0.521. These results highlight the effectiveness of our approach in accurately retrieving and integrating knowledge to solve complex questions.

\textbf{(2) PiCA shows strong generalization to out-of-domain scenarios}. Across five out-of-domain benchmarks, our method achieves the highest average EM and F1 scores for both model scales. Notably, PiCA leads in challenging multi-hop tasks like MuSiQue and Bamboogle, outperforming strong RL-based methods such as TIPS. This consistent advantage across TriviaQA and 2Wiki underscores the robustness of PiCA when encountering diverse, unseen data distributions.

\begin{table*}[!t]
\caption{F1 scores on seven QA benchmarks. In-domain tasks include NQ and HotpotQA; out-of-domain tasks include TriviaQA, PopQA, 2WikiMultiHopQA, MuSiQue, and Bamboogle.}
\centering
\small
\setlength{\tabcolsep}{6pt}
\begin{tabular}{l | cc | ccccc | c}
\toprule
\multirow{2}{*}{Method}
& \multicolumn{2}{c|}{In-domain}
& \multicolumn{5}{c|}{Out-of-domain}
& \multirow{2}{*}{Avg} \\
\cmidrule(lr){2-3}
\cmidrule(lr){4-8}
& NQ & HotpotQA
& TriviaQA & PopQA & 2Wiki & MuSiQue & Bamboogle
& \\
\midrule

\multicolumn{9}{c}{\textbf{Qwen2.5-3B Instruct}} \\
\midrule
RAG         & --    & 0.359 & --    & --    & 0.316 & 0.135 & 0.161 & -- \\
Search-o1   & --    & 0.326 & --    & --    & 0.309 & 0.117 & 0.436 & --\\
Search-R1   & 0.457 & 0.376 & 0.652 & 0.431 & 0.352 & 0.171 & 0.344 & 0.398 \\
ZeroSearch  & --    & 0.353 & --    & --    & 0.288 & 0.145 & 0.299 & -- \\
StepSearch  & --    & \underline{0.452} & --    & --    & \underline{0.385} & \textbf{0.261} & \underline{0.452} & -- \\
MT-PPO      & 0.478 & 0.342 & 0.629 & 0.448 & 0.261 & 0.111 & 0.140 & 0.344 \\
TIPS        & \underline{0.518} & 0.415 & 0.664 & \textbf{0.474} & 0.351 & 0.159 & 0.298 & \underline{0.411} \\
\midrule
\textbf{Ours}
& \textbf{0.521} & \textbf{0.514}
& \textbf{0.691} & \underline{0.469} & \textbf{0.472} & \underline{0.235} & \textbf{0.463}
& \textbf{0.481} \\

\midrule
\multicolumn{9}{c}{\textbf{Qwen2.5-7B Instruct}} \\
\midrule
RAG         & --    & 0.391 & --    & --    & 0.226 & 0.142 & 0.316 & -- \\
Search-o1   & --    & 0.288 & --    & --    & 0.289 & 0.127 & 0.427 & -- \\
Search-R1   & \underline{0.496} & 0.484 & 0.693 & 0.428 & 0.465 & 0.256 & 0.501 & 0.475 \\
ZeroSearch  & --    & 0.432 & --    & --    & 0.370 & 0.204 & 0.409 & -- \\
StepSearch  & --    & 0.502 & --    & --    & 0.431 & \textbf{0.312} & \textbf{0.530 }& -- \\
MT-PPO      & 0.506 & 0.353 & 0.621 & 0.465 & 0.286 & 0.121 & 0.209 & 0.366 \\
TIPS        & {0.483} & \underline{0.533} & \underline{0.700} & \underline{0.490} & \textbf{0.466} & 0.278 & \underline{0.516} & \underline{0.495} \\
\midrule
\textbf{Ours}
& \textbf{0.553} & \textbf{0.542}
& \textbf{0.713} & \textbf{0.493} & \textbf{0.466} & \underline{0.287} & \textbf{0.530}
& \textbf{0.512} \\

\bottomrule
\end{tabular}
\label{tab:main_results_f1}
\vspace{-12pt}
\end{table*}

\begin{figure}[!t]
    \centering
    \begin{subfigure}[b]{0.45\linewidth}
        \centering
        \includegraphics[width=\linewidth]{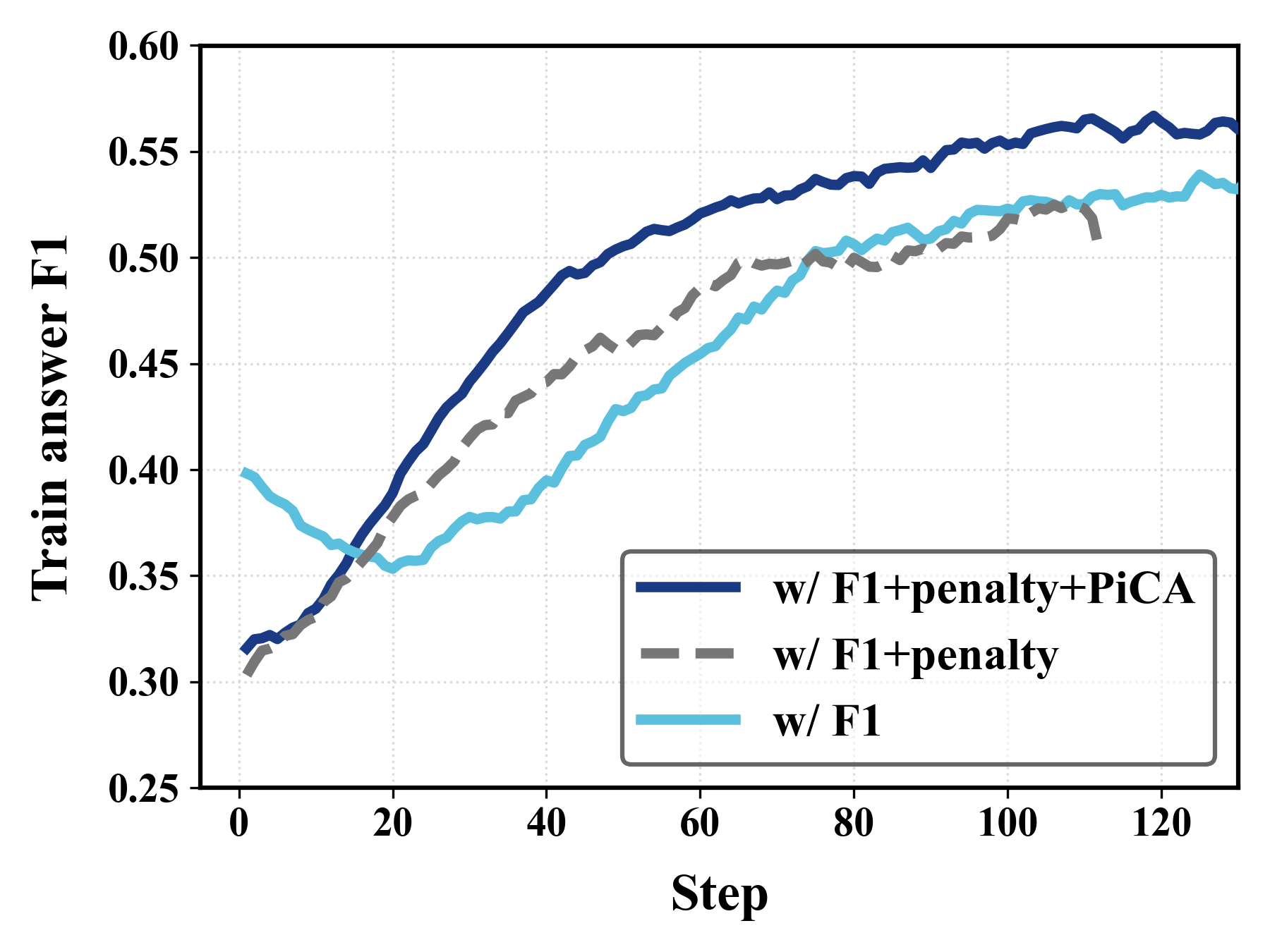}
        \caption{Answer F1 score}
        \label{fig:ablation_training_f1}
    \end{subfigure}
    \hspace{0.5em}
    \begin{subfigure}[b]{0.45\linewidth}
        \centering
        \includegraphics[width=\linewidth]{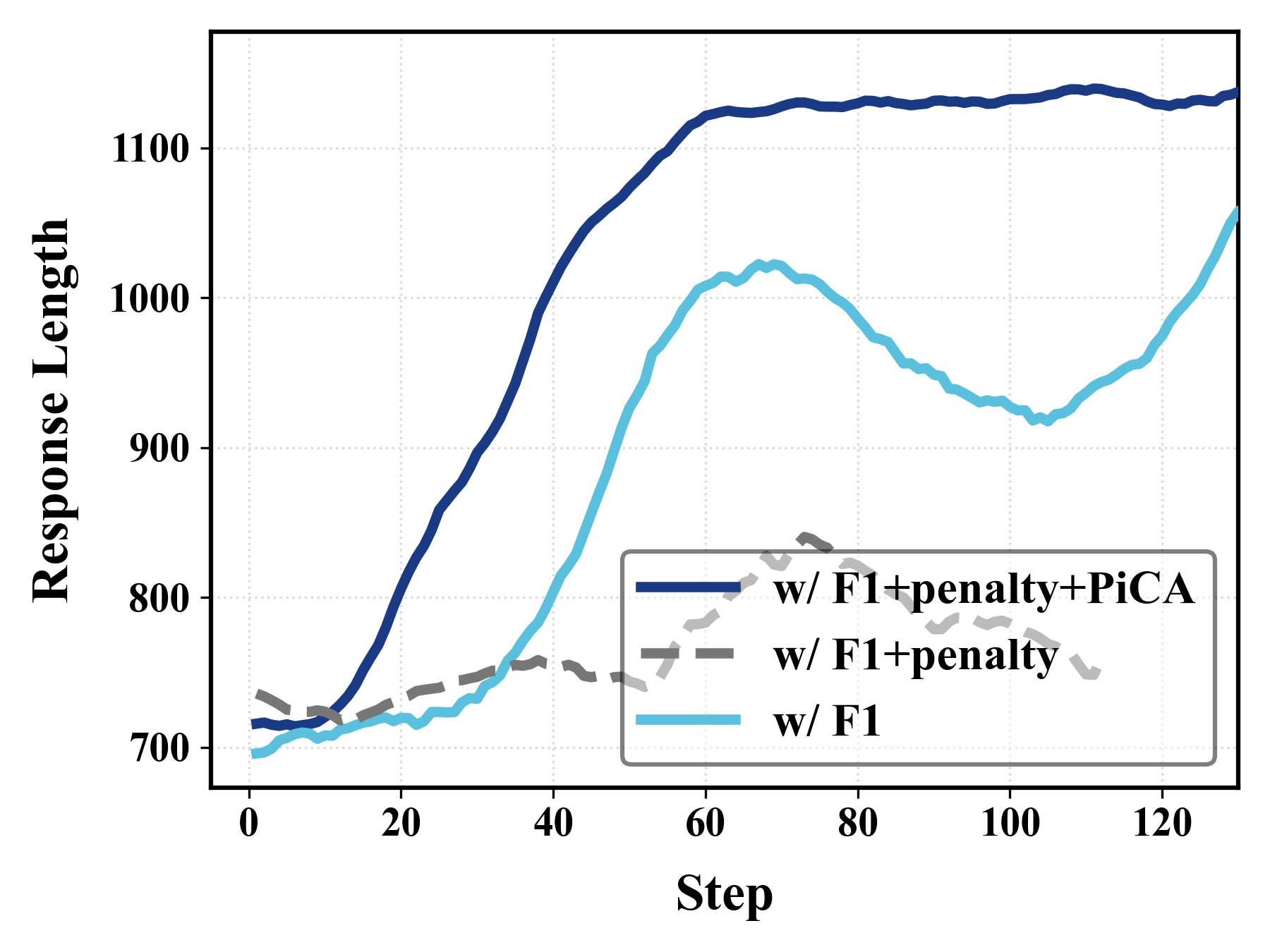}
        \caption{Response length}
        \label{fig:ablation_response_len}
    \end{subfigure}
    \caption{Comparison of PiCA with different rewards}
    \label{fig:ablation_fig}
    \vspace{-10pt}
\end{figure}

\vspace{-5pt}
\subsection{Ablation Study}\label{sec: ablation study}
\vspace{-5pt}

\begin{figure}[t]
    \centering
    \includegraphics[width=\linewidth, keepaspectratio]{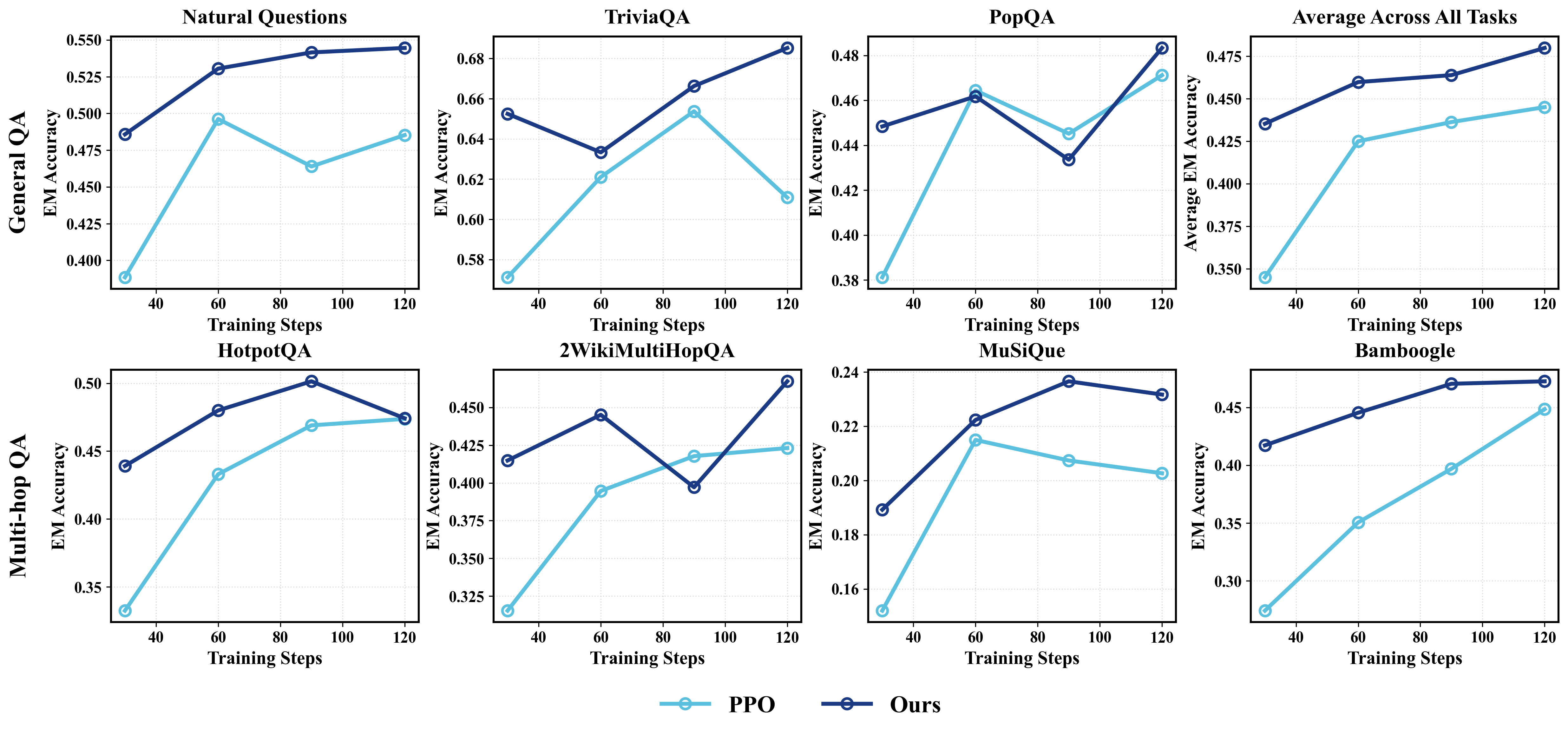}
    \caption{Evaluation results of PPO vs. PiCA}
    \label{fig:ablation_eval_results}
     \vspace{-10pt}
\end{figure}

To further investigate the contributions of PiCA, we conduct ablation experiments comparing three configurations: (w/ F1), which uses only rule-based outcome rewards; (w/ F1+penalty), which incorporates a negative reward for each search step to encourage efficiency; and (w/ F1+penalty+PiCA), our full hybrid reward framework as shown in Figure~\ref{fig:ablation_fig} and Figure~\ref{fig:ablation_eval_results}. We observe as follows:

\textbf{(1) PPO with our step reward demonstrates the most prominent performance advantages.}
Our reward consistently outperforms standard PPO across all benchmarks, with evaluation performance closely mirroring training gains. Unlike standard PPO, our method maintains a steady upward trajectory, ensuring superior reasoning accuracy and long-term optimization stability.

\textbf{(2) Step penalty rewards significantly accelerate convergence and our rewards can prevent training collapse.}
Before 40 steps, penalties rewards accelerate early convergence by discouraging inefficient, long reasoning paths that typically yield negative outcomes. However, relying solely on penalties eventually triggers length hacking, where the model curtails response length to minimize costs in Figure~\ref{fig:ablation_response_len}. As shown in Figure~\ref{fig:ablation_training_f1}, this behavior leads to the performance collapse observed in the w/ F1+penalty model after step 100. In contrast, PiCA maintains stable response lengths by balancing penalties with learned rewards.

\vspace{-5pt}
\subsection{More Analysis}
\begin{figure}[t] 
    \centering
    
    \begin{minipage}[c]{0.55\textwidth}
        \centering
        \captionof{table}{Generalization of PiCA across model families and scales. EM/F1 are PiCA scores; percentages in parentheses indicate relative improvement over the outcome-only PPO baseline.} 
        \label{tab:more_generalization}
        
        \resizebox{\linewidth}{!}{ 
            \begin{tabular}{lcc}
                \toprule
                Model & EM & F1  \\
                \midrule
                Qwen2.5-3B-Instruct      & 39.6  (+15.1\%) & 48.1  (+13.7\%)  \\
                Qwen3-4B-Instruct-2507   & 45.0  (+6.6\%)  & 54.3  (+4.1\%)   \\
                Qwen2.5-7B-Instruct      & 42.6  (+12.1\%) & 51.2  (+12.9\%)  \\
                Llama3.1-8B-Instruct     & 40.2  (+34.0\%) & 48.5  (+29.1\%)  \\
                \bottomrule
            \end{tabular}
        }
    \end{minipage}
    \hfill 
    \begin{minipage}[c]{0.43\textwidth}
        \centering
      
        \includegraphics[width=\linewidth]{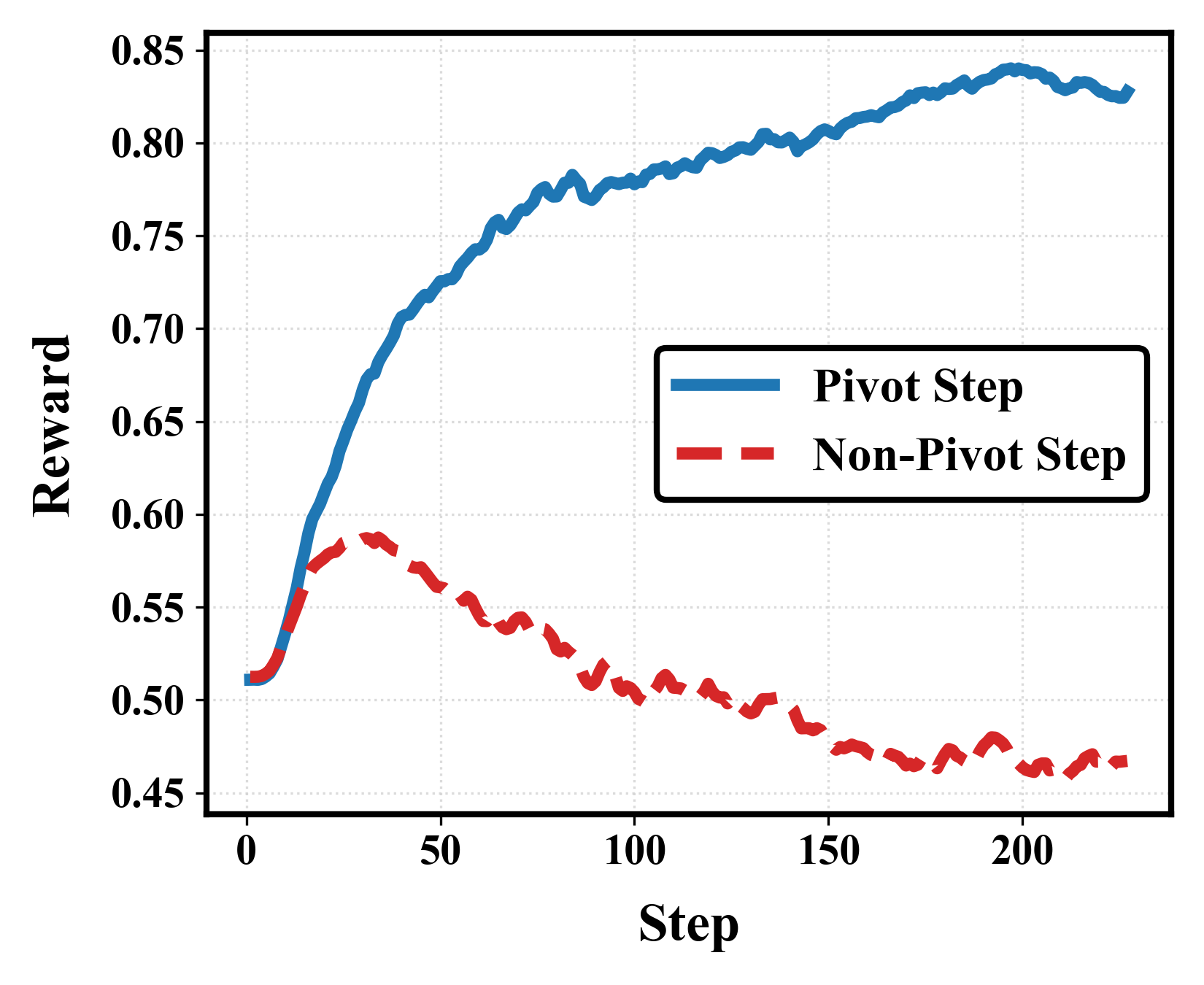}
        \caption{{PiCA step rewards.}}
        \label{fig:step_reward}
    \end{minipage}
    
    \vspace{-10pt} 
\end{figure}

\vspace{-5pt}
\textbf{Generalization of PiCA}. To evaluate the generalization of our method, we apply the same training setup to various base models across different families and scales. As shown in Table~\ref{tab:more_generalization}, our framework consistently enhances performance in all cases. While the 3B-scale model (Qwen2.5-3B) exhibits the most significant relative growth, the improvements remain consistent even on stronger base models. For instance, Qwen3-4B, which possesses higher initial capabilities, still achieves a +6.6\% EM and +4.1\% F1 boost. Furthermore, the method scales effectively to larger architectures like Qwen2.5-7B (+12.1\%/12.9\%) and Llama3.1-8B (+34.0\%/29.1\%), indicating that our framework robustly enhances search capability across different model families and parameter scales.

\textbf{Step Reward Showcase}. To evaluate the precision of PiCA, we conduct a fine-grained analysis of reward distribution during reward model training. As shown in Figure~\ref{fig:step_reward}, PiCA effectively discriminates between \textbf{Pivot} and \textbf{Non-Pivot Steps}, with rewards diverging toward $0.8$ and $0.45$, respectively. Though normalized to $[0, 1]$ for visualization, these values correspond directly to the $[-1, 1]$ range used in our optimization framework. This provides a robust signal that prioritizes high-information-gain reasoning steps, as further detailed in the qualitative examples in Appendix~\ref{app:case_study}.

 \vspace{-9pt}
\section{Conclusion}
 \vspace{-9pt}
In this work, we present Pivot-Based Credit Assignment (PiCA), a novel credit assignment designed for search agentic reinforcement learning in knowledge-intensive tasks. By reformulating the search trajectory as a sequential process of cumulative search progress, PiCA effectively mitigates critical challenges in long-horizon credit assignment, such as reward sparsity and isolated credit. Our PiCA is theoretically grounded in Potential-Based Reward Shaping (PBRS) and identifies pivot steps as information peaks to provide dense, trajectory-dependent guidance without the risk of distributional shift. Extensive experiments over seven diverse multi-hop QA benchmarks demonstrate that PiCA achieves state-of-the-art performance, outperforming competitive baselines by up to 15.2\% while ensuring the retrieval of verifiable and accurate evidence.
\clearpage
\begin{ack}
This work was supported by the National Natural Science Foundation of China under Grant No. 62572418.

\end{ack}

{
\bibliographystyle{plain}
\bibliography{references}
}

\clearpage
\input{checklist.tex}

\clearpage
\appendix

\section{Data Generation}\label{app:generation}
\textbf{Process}. As described in introduction about \textbf{pivot steps}, we enrich approximately 12,000  training instances from the StepSearch~\cite{stepsearch} dataset. For each question, we interact with real search environment and utilize the DeepSeek-V3 API ~\cite{deepseekai2025deepseekv3technicalreport} to sample 5 times. Each trajectory is then annotated using an LLM-as-a-judge approach~\cite{gu2025surveyllmasajudge}, following the structured evaluation prompts detailed in Appendix~\ref{app:generation}. Through a sequential matching process, an intermediate step is labeled as a positive pivot step ($+$) if it successfully gives a target golden sub-answer with its corresponding query, while all other steps are assigned a negative label ($-$). Formally, we define $\mathcal{D}_p \subseteq \mathcal{D}$ as the set of all pivot steps. In addition to step-level signals, we assign binary outcome labels $l \in \{0, 1\}$ by performing an Exact Match (EM) between the final generated answer and the ground truth. To ensure the high fidelity of the training data, we implement a rigorous filtering pipeline that discards trajectories with structural inconsistencies, such as mismatches between the number of generated search rounds and their corresponding step labels. Furthermore, we conduct human-in-the-loop audits on a random subset of the data to verify that the identified pivot steps represent genuine logical junctures in the reasoning process.

\textbf{Data Quality}.
To assess the quality of the generated data, we randomly sample 100 test trajectories and have 5 human annotators audit the pivot labels. Since fully annotating the entire dataset is infeasible, we reuse the 100 annotated trajectories and find that approximately 5\% of the labels in the original training set are noisy; despite this, PiCA still achieves performance improvement. To further test robustness, we conduct an additional noise injection experiment by randomly flipping 10\% of the DeepSeek-V3 annotated labels. We then examine distributional consistency by using the 100-step policy model to generate rollouts on the same 100 test queries, checking whether the reward model remains consistent with human annotations as the policy evolves. All metrics (\emph{i.e.}, Human Judge Accuracy, Precision, and Recall) are computed against the majority-vote labels from the five annotators. The results in Table~\ref{tab:data_quality} demonstrate that PiCA remains robust under 10\% label noise and stays aligned with human judgments throughout policy evolution.

\begin{table*}[h]
\centering
\caption{Evaluation of data quality under default, 10\% label noise, and 100-step training settings, measured by human judge accuracy, precision, and recall.}
\label{tab:data_quality}
\begin{tabular}{lccc}
\toprule
Setting & Human Judge Accuracy & Precision & Recall \\
\midrule
Default & 0.9160 & 0.8357 & 0.8719 \\
10\% Label Noise & 0.8530 & 0.8013 & 0.8526 \\
100-step Training & 0.9050 & 0.8270 & 0.8651 \\
\bottomrule
\end{tabular}
\end{table*}

\textbf{Prompt}. The prompts below are used to generate step label for our reward model training. The labeling process evaluates each step in a trajectory against the golden sub-queries and sub-answers from the MuSiQue dataset in StepSearch. A step is labeled '+' if its search and reasoning successfully yield the corresponding golden sub-answer; otherwise, it receives a '-' label.
\begin{tcolorbox}[colback=white,colframe=gray!50!black,title=Trajectory Evaluation Prompt, breakable,]
As a Search-QA auditor, your task is to evaluate a multi-hop reasoning trajectory. 
You must compare each step's <search> and <think> process against the provided Golden Reference.
\newline
[Standard Reference]
\newline
- Ground Truth (GT): $\{$gt$\}$\
\newline
- Golden Min Hop: $\{$hop$\}$
\newline
- Golden Sub-Queries: $\{$golden\_queries$\}$
\newline
- Golden Sub-Answers: $\{$golden\_answers$\}$
\newline
[Trajectory for Evaluation]
$\{$clean\_trajectory$\}$
\newline
[Strict Evaluation Rules]
\newline
Assign '+' or '-' based on these logic checks:\newline
1. Step Quality (+):\newline
   - The <search> query effectively matches the intent of the 'Golden Sub-Queries' for that step.\newline
   - If the <information> block is missing the 'Golden Sub-Answers', but the model correctly recognizes this in <think> and searches again, it is still '+'.\newline
2. Step Failure (-):\newline
   - Summary Omission: The <information> block contains the 'Golden Sub-Answers', but the model's <think> block ignores it or fails to summarize it, leading to a redundant or wrong next step.\newline
   - Logic Deviation: The <search> query significantly departs from the 'Golden Sub-Queries' logic or targets the wrong entity.\newline
   - Hallucination: The <think> block claims to have found information that is not in the <information> block.\newline
   - Fact Mismatch: The step produced an answer that contradicts or deviates from the Golden Sub-Answers/GT (e.g., searching for a location and picking the wrong entity from the results).\newline
   - False Positive Search: The query looks correct, but the model extracted the WRONG information from the search results.\newline
   - Final Answer Conflict: The final answer is {gt} but the model produced {ans}. This implies at least one step in the trajectory MUST have a logical flaw or information extraction error. Identify it and mark it '-'.\newline
[Output Requirement]\newline
Return ONLY a JSON object with:\newline
"reason": A brief explanation of where the logic failed.\newline
"labels": An array of '+' or '-' corresponding to each `</search>` tag in order.\newline
Example: {{"reason": "Step 2 found the info but the model didn't summarize it.", "labels": ["+", "-", ...]}}
\newline
Note: The length of the "labels" array MUST exactly match the number of `</search>` tags.
\end{tcolorbox}

\begin{tcolorbox}[colback=white,colframe=gray!50!black,title=Data Example, breakable,]
  "question": "When did the alma mater of William C. Perry start issuing engineering degrees?",\newline
  "trajectory": [
    $\{$
    \newline
      "step": 1,
      "observation": None,
      "think": <think>I need to parse the question: "When did the alma mater of William C. Perry start issuing engineering degrees?" First, I need to identify who William C. Perry is. Likely William C. Perry is a notable person, possibly a former US Secretary of Defense (William Perry, served 1994-1997). His full name is William James Perry, but sometimes referred to as William C. Perry? Actually, William Perry's middle name is James, not C. Maybe it's a different William C. Perry. Could be William C. Perry, maybe an academic or engineer. Let's search for William C. Perry to confirm. </think>
      "action": <search>William C. Perry alma mater</search>,
      "is\_pivot": true,
     $\}$,
    \newline
    $\{$
      "step": 2,
      "observation": <information>Doc 1 (Title: William C. Perry)... Doc 2 (Title: Perry Belmont)... Doc 3 (Title: William Perry (American football))... </information>,
      "think": <think>The search results show a William C. Perry who graduated from the University of Kansas in 1922. That's likely the person in question. So his alma mater is the University of Kansas. The question asks: "When did the alma mater of William C. Perry start issuing engineering degrees?" So we need to find when the University of Kansas started issuing engineering degrees. We need to search for the history of engineering degrees at the University of Kansas. </think>
      "action": <search>University of Kansas start issuing engineering degrees</search>,
      "is\_pivot": true,
     $\}$,
    \newline
    $\{$
      "step": 3,
      "observation": <information>Doc 1 (Title: University of Kansas)... Doc 2 (Title: University of Kansas)... Doc 3 (Title: University of Kansas)... </information>,
      "think": None
      "action": <answer>1873</answer>,
      "is\_pivot": true,
      
    $\}$
  ],
   \newline
  "final\_answer": "1873"
\end{tcolorbox}


\section{Generalize to web scenario and long-horizon scenario}\label{app:benchmark}

\textbf{Web Scenario}. We further evaluate PiCA in a real-world web search scenario to assess its generalization ability,
with results reported in Table~\ref{tab:web}. Following ARPO~\cite{dong2025agenticreinforcedpolicyoptimization}, we evaluate our methods on GAIA~\cite{mialon2023gaiabenchmarkgeneralai}, Webwalker~\cite{wu2025webwalkerbenchmarkingllmsweb} and Humanity’s Last Exam (HLE)~\cite{2026hle} benchmarks. We report the Avg@3 metric which averages the F1 scores and LLM-as-a-Judge scores across
3 independent runs. During evaluation, each
search retrieves up to 3 web pages, which are then
processed by a summarization model instantiated
with Qwen2.5-7B-Instruct.
On average across GAIA, Webwalker, and HLE benchmarks, PiCA outperforms all baselines, demonstrating strong generalization to real-world web search tasks.

\textbf{Long-horizon Scenario}. To assess generalization in long-horizon search tasks, we additionally train Qwen2.5-7B-Instruct with 10 retrieval turns. Table~\ref{tab:long-horizon} reports the results on NQ, TriviaQA, PopQA, HotpotQA, 2Wiki, MuSiQue, and Bamboogle. The results demonstrate that PiCA consistently improves performance over PPO and generalizes to longer-horizon search.
\begin{table*}[t]
\centering
\caption{Avg@3 performance on GAIA, Webwalker and HLE with real web search engine. The best results are indicated with \textbf{bold}.}
\label{tab:web}
\begin{tabular}{lcccccc}
\toprule
Benchmark & Search-R1 & Stepsearch & Zerosearch & TIPS & IGPO & PiCA \\
\midrule
GAIA  & 18.45 & 12.23 & 16.50 & 19.22 &  \underline{19.75} & \textbf{20.44} \\
Webwalker  & \textbf{13.99} & 10.50 & 13.11 & 12.15 & 13.64 &  \underline{13.97} \\
Humanity's Last Exam (HLE)  & 6.17 & 4.66 & 4.51 & 5.64 & \textbf{7.02} & \underline{6.46} \\
\bottomrule
\end{tabular}
\end{table*}

\begin{table*}[t]
\centering
\caption{Exact Match results on seven QA benchmarks. In-domain tasks include NQ and HotpotQA; out-of-domain tasks include TriviaQA, PopQA, 2WikiMultiHopQA, MuSiQue, and Bamboogle.}
\label{tab:long-horizon}
\begin{tabular}{lccccccc}
\toprule
Method & NQ & TriviaQA & PopQA & HotpotQA & 2Wiki & MuSiQue & Bamboogle \\
\midrule
PPO  & 0.455 & 0.663 & 0.437 & 0.418 & 0.402 & 0.199 & 0.432 \\
PiCA & 0.469 & 0.711 & 0.465 & 0.430 & 0.427 & 0.213 & 0.472 \\
\bottomrule
\end{tabular}
\end{table*}

\textbf{Pivot Step Distribution Analysis}.  For both QA and deepsearch task, we use Qwen2.5-7B-Instruct to rollout trajectories. Then, we use PiCA Qwen2.5-3B-Instruct reward model to score each turn. Here turn1 represents the proportion of pivots within turn1 data. The result in Table~\ref{tab:count} shows that DeepResearch tasks face more complex environments and need multi-hop reasoning, so they need longer trajectories, while QA tasks are short-hop questions with a specific environment.
Specifically, DeepSearch maintains a relatively high pivot ratio in later turns and peaks at turn 7 (0.8254), suggesting that long-horizon tasks require continued strategic re-planning beyond the early turns. In contrast, QA peaks at turn 3 (0.8317) and then drops at turn 4 (0.7364), consistent with its shorter trajectory and limited need for iterative reasoning.

\begin{table}[h]
\centering
\caption{Pivot step distribution across turns for QA and DeepSearch tasks.}
\label{tab:count}
\begin{tabular}{lcc}
\toprule
Turn & QA task & Deepsearch task\\
\midrule
Turn Counts & 4.72 & 7.18 \\
turn0 & 0.7620 & 0.7413 \\
turn1 & 0.7689 & 0.7796 \\
turn2 & 0.8020 & 0.8175 \\
turn3 & 0.8317 & 0.8237 \\
turn4 & 0.7364 & 0.7792 \\
turn5 & --     & 0.7837 \\
turn6 & --     & 0.7931 \\
turn7 & --     & 0.8254 \\
turn8 & --     & 0.7675 \\
turn9 & --     & 0.7282 \\
\bottomrule
\end{tabular}
\end{table}

\section{Prompt for Research Plan on Question Answering}
Following Search-R1, the prompts below are used to generate search trajectories during policy optimization.
\begin{tcolorbox}[colback=white,colframe=gray!50!black,title=Trajectory Generation Prompt, breakable,]
Answer the given question. You must conduct reasoning inside \tokenthink{think} and \tokenthink{/think} first every time you get new information. After reasoning, if you find you lack some knowledge, you can call a search engine by \tokensearch{search} query \tokensearch{/search} and it will return the top searched results between \tokeninfo{information} and \tokeninfo{/information}. You can search as many times as your want. If you find no further external knowledge needed, you can directly provide the answer inside \tokenanswer{answer} and \tokenanswer{/answer}, without detailed illustrations. For example, <answer> Beijing </answer>. Question: $\{$question$\}$\
\end{tcolorbox}

\section{Details of Benchmarks and Metrics}
\label{app:datasets}

\subsection{General Question Answering}

\paragraph{Natural Questions (NQ).}
Natural Questions (NQ)~\cite{kwiatkowski2019natural} is a large-scale QA benchmark based on real Google search queries, where annotators label long answers (paragraph-level) and, when possible, short spans or yes/no answers from Wikipedia pages. The dataset contains approximately 307K training, 8K development, and 8K test examples, with annotations reflecting natural user information needs.We follow the standard evaluation protocol and report EM and F1 scores.

\paragraph{TriviaQA.}
TriviaQA~\cite{joshi2017triviaqa} is a large-scale open-domain QA dataset constructed from trivia and quiz sources, featuring diverse question formulations and multiple evidence documents per query. It contains over 95K question–answer pairs and more than 650K question–answer–evidence triples, with evidence drawn from both web pages and Wikipedia. This setup challenges models in both retrieval and reasoning. We follow prior work and report EM and F1.

\paragraph{PopQA.}
PopQA~\cite{mallen2023popqa} is an entity-centric QA benchmark derived from Wikidata triples, designed to evaluate performance across both popular and long-tail knowledge. It contains approximately 14K examples, each constructed from subject–relation–object triples and enriched with metadata such as entity IDs, relation types, and Wikipedia page-view statistics. This setup enables analysis of retrieval bias and factual memorization. We report EM and F1 following standard evaluation.

\subsection{Multi-Hop Question Answering}

\paragraph{HotpotQA.}
HotpotQA~\cite{yang2018hotpotqa} is a multi-hop QA dataset requiring reasoning across multiple Wikipedia documents, with sentence-level supporting fact annotations to encourage explainable reasoning. It contains approximately 113K examples and supports both distractor and fullwiki settings. We evaluate performance using EM and F1.

\paragraph{2WikiMultiHopQA.}
2WikiMultiHopQA~\cite{ho2020twowiki} is a large-scale multi-hop QA benchmark combining Wikipedia text with structured knowledge from Wikidata. It includes around 192K examples and provides both supporting facts and explicit reasoning paths in the form of triples. We follow the standard evaluation protocol and report EM and F1.

\paragraph{MuSiQue.}
MuSiQue~\cite{trivedi2022musique} constructs multi-hop questions by composing independent single-hop questions, enforcing compositional reasoning and reducing shortcut learning. The dataset contains about 25K questions spanning 2–4 hops and provides intermediate reasoning steps. We report EM and F1 following prior work.

\paragraph{Bamboogle.}
Bamboogle~\cite{press2023bamboogle} is a small but challenging dataset of 125 manually curated two-hop questions designed to minimize shortcut reasoning. Each question requires combining multiple facts from Wikipedia, while supporting evidence is not explicitly provided. We evaluate performance using EM and F1.

\subsection{Exact Match (EM)}
The Exact Match (EM) metric evaluates whether the predicted answer exactly matches any of the reference answers. Formally, EM is defined as a binary indicator:
\begin{equation}
\mathrm{EM}(\hat{y}, \mathcal{Y}) =
\begin{cases}
1, & \text{if } \exists\, y \in \mathcal{Y} \text{ such that } \hat{y} = y, \\
0, & \text{otherwise.}
\end{cases}
\end{equation}
Here, $\hat{y}$ denotes the predicted answer, and $\mathcal{Y}$ represents the set of all acceptable ground-truth answers.

\subsection{F1 Score}
The F1 score measures the token-level overlap between the predicted answer and a reference answer. Given a predicted token set $T_{\hat{y}}$ and a ground-truth token set $T_{y}$, the F1 score is computed as:
\begin{equation}
\mathrm{F1}(T_{\hat{y}}, T_{y}) = 
\frac{2 \cdot |T_{\hat{y}} \cap T_{y}|}
{|T_{\hat{y}}| + |T_{y}|}.
\end{equation}

When multiple reference answers are available, we compute the F1 score against each candidate and report the maximum value:
\begin{equation}
\mathrm{F1}(\hat{y}, \mathcal{Y}) = \max_{y \in \mathcal{Y}} \mathrm{F1}(T_{\hat{y}}, T_{y}).
\end{equation}

\section{Limitation and  Future Direction}
While our method demonstrates significant improvements across various QA tasks, due to limited computational resources, we are unable to train larger models (e.g., 14B and 32B series). Although the observed trends are consistent, future work could extend this validation to ultra-large-scale models or a broader diversity of architectures to ensure the approach's generalizability across varying model capacities.

Moreover, our current framework predominantly relies on process reward models that necessitate external supervision or high-quality teacher signals. In follow-up studies, we plan to investigate more autonomous reward mechanisms, such as intrinsic motivation, self-reflective feedback, or information-gain-based rewards. Such explorations would reduce the dependency on labor-intensive annotations and pave the way for a more self-evolving and robust agent.
\section{Broader Impacts}
By introducing a unified process reward framework, our work establishes a new paradigm for enhancing the transparency and reliability of LLM-based search agents. The transition from outcome-only supervision to granular, context-aware credit assignment not only mitigates the "black-box" nature of long-horizon reasoning but also ensures that agent behaviors are intrinsically aligned with verifiable information gain.

Furthermore, our approach provides a scalable solution for developing intelligent AI in knowledge-intensive domains. By effectively neutralizing the distribution inconsistency and credit assignment gaps, this research serves as a catalyst for the deployment of autonomous agents in high-stakes environments~(\emph{e.g.}, scientific research, legal analysis) where step-by-step accountability is paramount. Ultimately, by fostering more knowledgeable and self-correcting agents, our findings contribute to the broader goal of building verifiable, and human-aligned artificial intelligence systems that can navigate increasingly complex information landscapes.


\section{Implementation Details}\label{app:implementation}

\subsection{Hyper-parameters}
In Qwen backbone, we use <|vision\_start|> token as special token to give turn-level rewards.

\begin{table}[H]
\caption{Hyperparameter settings for RL training.}
\centering
\small
\setlength{\tabcolsep}{10pt}
\begin{tabular}{ll}
\toprule
\textbf{Parameter} & \textbf{Value} \\
\midrule
data.train\_batch\_size & 256 \\
data.val\_batch\_size & 128 \\
data.max\_prompt\_length & 8192 \\
data.max\_response\_length & 800 \\
data.max\_start\_length & 2048 \\
data.max\_obs\_length & 800 \\
data.shuffle\_train\_dataloader & True \\
algorithm.adv\_estimator & gae \\
algorithm.gamma & 1.0 \\
algorithm.kl\_ctrl.kl\_coef & 0.001 \\
algorithm.no\_think\_rl & False \\
actor\_rollout\_ref.actor.optim.lr & $7\mathrm{e}{-7}$ \\
actor\_rollout\_ref.actor.optim.lr\_warmup\_steps\_ratio & 0.285 \\
actor\_rollout\_ref.actor.ppo\_mini\_batch\_size & 32 \\
actor\_rollout\_ref.actor.ppo\_micro\_batch\_size & 8 \\
actor\_rollout\_ref.actor.state\_masking & True \\
actor\_rollout\_ref.model.enable\_gradient\_checkpointing & True \\
actor\_rollout\_ref.model.use\_remove\_padding & True \\
actor\_rollout\_ref.actor.fsdp\_config.param\_offload & True \\
actor\_rollout\_ref.actor.fsdp\_config.grad\_offload & True \\
actor\_rollout\_ref.actor.fsdp\_config.optimizer\_offload & True \\
actor\_rollout\_ref.rollout.name & vllm \\
actor\_rollout\_ref.rollout.temperature & 1.0 \\
actor\_rollout\_ref.rollout.n\_agent & 5 \\
actor\_rollout\_ref.rollout.tensor\_model\_parallel\_size & 1 \\
actor\_rollout\_ref.rollout.gpu\_memory\_utilization & 0.6 \\
actor\_rollout\_ref.rollout.log\_prob\_micro\_batch\_size & 16 \\
actor\_rollout\_ref.ref.log\_prob\_micro\_batch\_size & 16 \\
actor\_rollout\_ref.ref.fsdp\_config.param\_offload & True \\
critic.optim.lr & $7\mathrm{e}{-6}$ \\
critic.optim.lr\_warmup\_steps\_ratio & 0.015 \\
critic.ppo\_micro\_batch\_size & 16 \\
critic.model.enable\_gradient\_checkpointing & True \\
critic.model.use\_remove\_padding & True \\
critic.model.fsdp\_config.param\_offload & True \\
critic.model.fsdp\_config.grad\_offload & True \\
critic.model.fsdp\_config.optimizer\_offload & True \\
reward\_model.url & localhost:5000/get\_reward \\
step\_reward\_scale & 0.3 \\
baseline\_step\_reward & 0.55 \\
outcome\_reward\_scale & 1.5 \\
trainer.critic\_warmup & 0 \\
trainer.n\_gpus\_per\_node & 8 \\
trainer.nnodes & 1 \\
trainer.total\_epochs & 15 \\
trainer.total\_training\_steps & 190 \\
trainer.save\_freq & 60 \\
trainer.test\_freq & 30 \\
max\_turns & 5 \\
retriever.topk & 3 \\
retriever.url & localhost:8000/retrieve \\
\bottomrule
\end{tabular}
\label{tab:rl_hyperparams}
\end{table}

\begin{table}[H]
\caption{Hyperparameter settings for Reward Model.}
\centering
\small
\setlength{\tabcolsep}{6pt}
\begin{tabular}{l l}
\toprule
\textbf{Parameter} & \textbf{Value} \\
\midrule

model.pretrain & 3B model \\
data.split (train/test) & train / test \\
data.input\_key & input \\
data.label\_key & value \\

attn.implementation & flash\_attention\_2 \\
training.packing & True \\
training.placeholder\_token & \texttt{<|vision\_start|>} \\
reward.tokens & \{+, -\} \\
crm.gamma & 1.0 \\

sequence.max\_length & 32768 \\
batch.train\_batch\_size & 256 \\
batch.micro\_batch\_size & 16 \\

optimization.learning\_rate & $5 \times 10^{-6}$ \\
optimization.max\_epochs & 1 \\
optimization.max\_samples & 100K \\
precision & bf16 \\

parallel.zero\_stage & 3 \\
training.gradient\_checkpointing & True \\
training.num\_gpus & 2 \\

logging.save\_steps & 500 \\
logging.eval\_steps & 100 \\
logging.logging\_steps & 1 \\

\bottomrule
\end{tabular}
\label{tab:crm_config}

\vspace{-15pt}
\end{table}

\section{Case Study}\label{app:case_study}

\begin{figure}[H]
    \centering
    \includegraphics[width=\linewidth]{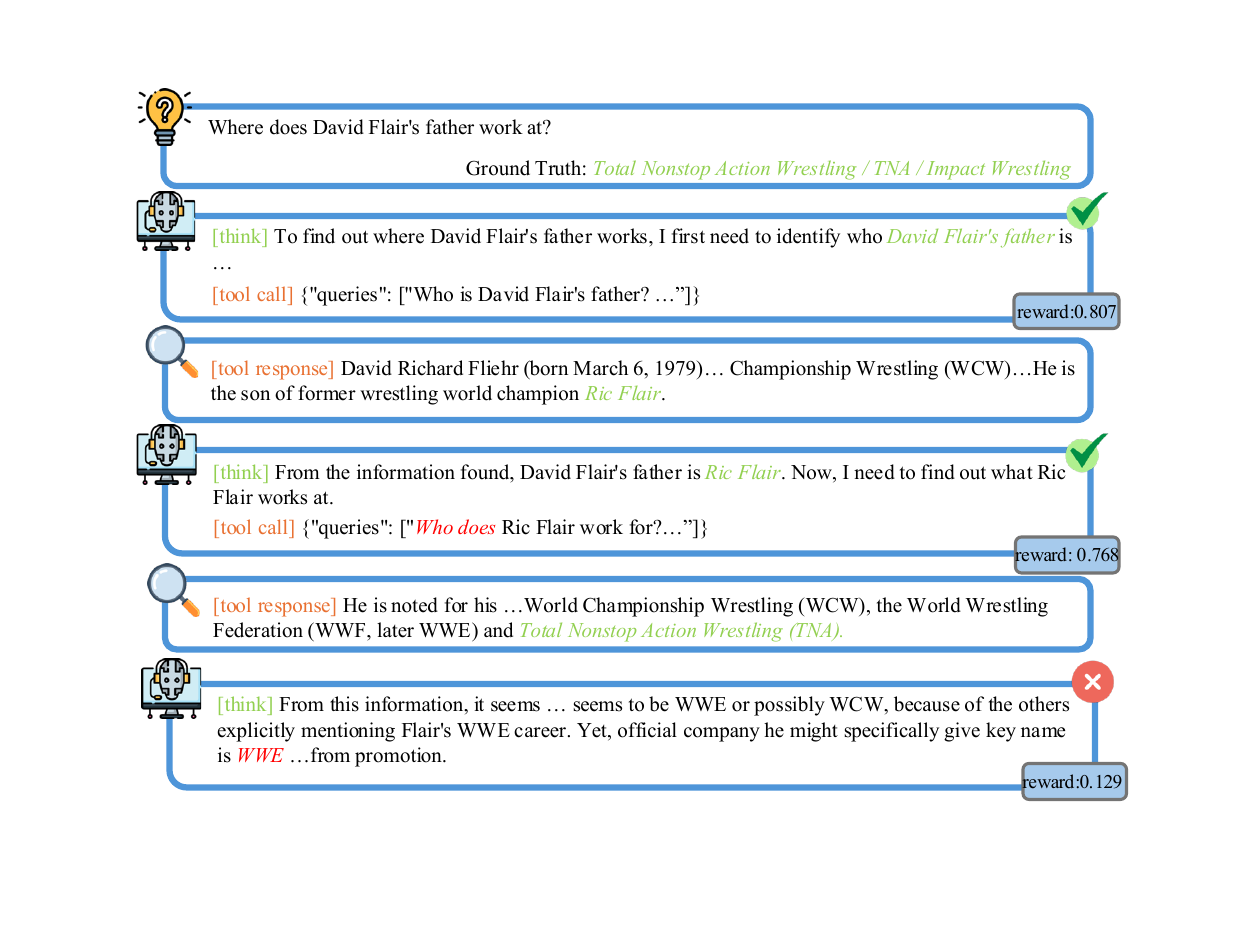}
    \caption{this case study presents a failed trajectory with an incorrect final answer. Although the overall result is wrong, some intermediate steps still align with the golden reasoning process. PiCA assigns high positive rewards to these informative steps, while giving lower or negative rewards to redundant or misleading steps that lead to the incorrect answer.}
    \label{fig:case_study_wrong}
\end{figure}

\begin{figure}[t]
    \centering
    \includegraphics[width=\linewidth]{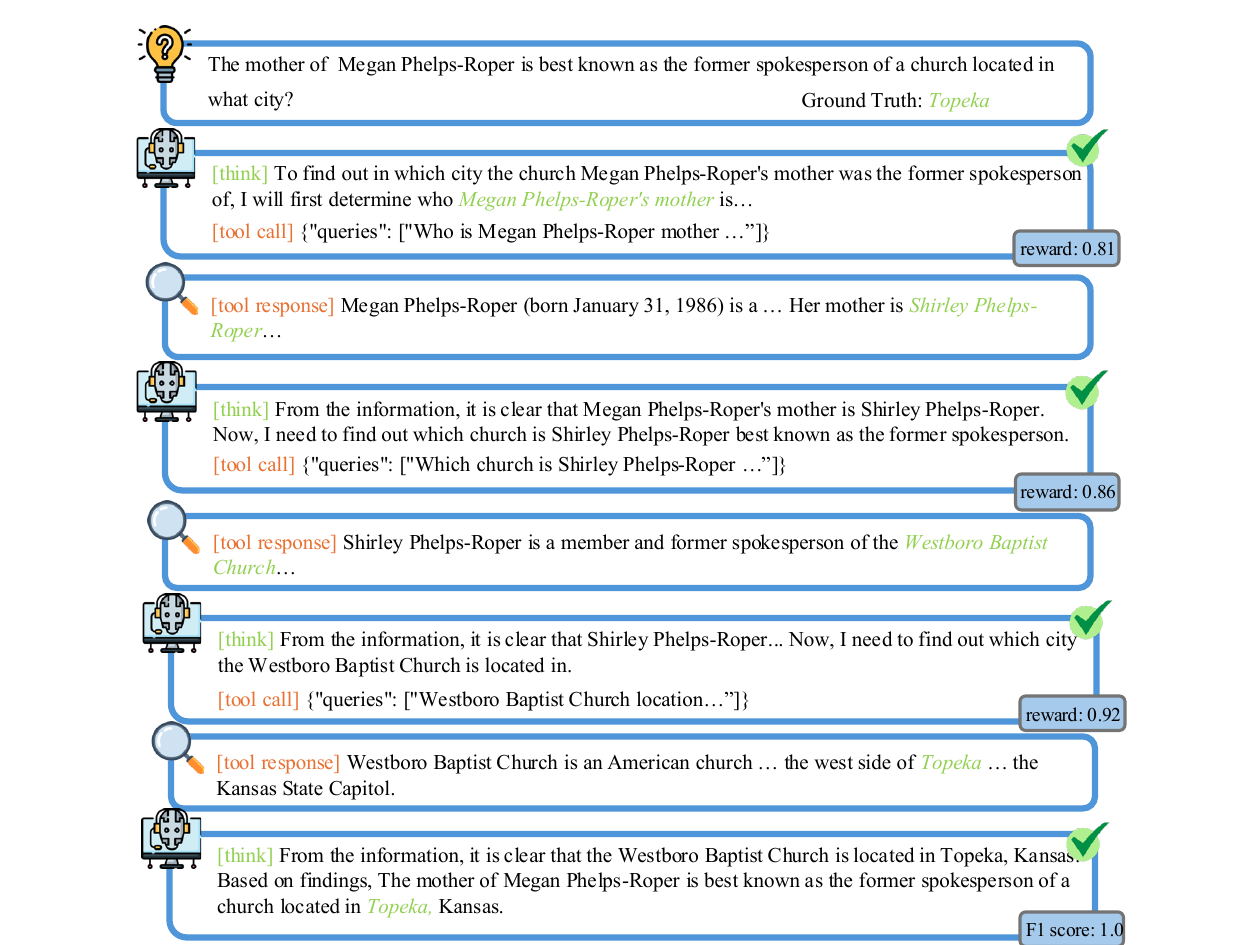}
    \caption{this case study presents a successful reasoning trajectory in which the intermediate reasoning and retrieval steps consistently align with the golden process, leading to a correct final answer. In such cases, PiCA assigns high positive rewards to each informative step, reflecting the consistency between the reasoning trajectory and the target solution path.}
    \label{fig:case_study_right}
\end{figure}



\end{document}

%% file: checklist.tex
\section*{NeurIPS Paper Checklist}

The checklist is designed to encourage best practices for responsible machine learning research, addressing issues of reproducibility, transparency, research ethics, and societal impact. Do not remove the checklist: {\bf The papers not including the checklist will be desk rejected.} The checklist should follow the references and follow the (optional) supplemental material.  The checklist does NOT count towards the page
limit. 

Please read the checklist guidelines carefully for information on how to answer these questions. For each question in the checklist:
\begin{itemize}
    \item You should answer \answerYes{}, \answerNo{}, or \answerNA{}.
    \item \answerNA{} means either that the question is Not Applicable for that particular paper or the relevant information is Not Available.
    \item Please provide a short (1--2 sentence) justification right after your answer (even for \answerNA). 
\end{itemize}

{\bf The checklist answers are an integral part of your paper submission.} They are visible to the reviewers, area chairs, senior area chairs, and ethics reviewers. You will also be asked to include it (after eventual revisions) with the final version of your paper, and its final version will be published with the paper.

The reviewers of your paper will be asked to use the checklist as one of the factors in their evaluation. While \answerYes{} is generally preferable to \answerNo{}, it is perfectly acceptable to answer \answerNo{} provided a proper justification is given (e.g., error bars are not reported because it would be too computationally expensive'' or ``we were unable to find the license for the dataset we used''). In general, answering \answerNo{} or \answerNA{} is not grounds for rejection. While the questions are phrased in a binary way, we acknowledge that the true answer is often more nuanced, so please just use your best judgment and write a justification to elaborate. All supporting evidence can appear either in the main paper or the supplemental material, provided in appendix. If you answer \answerYes{} to a question, in the justification please point to the section(s) where related material for the question can be found.

IMPORTANT, please:
\begin{itemize}
    \item {\bf Delete this instruction block, but keep the section heading ``NeurIPS Paper Checklist"},
    \item  {\bf Keep the checklist subsection headings, questions/answers and guidelines below.}
    \item {\bf Do not modify the questions and only use the provided macros for your answers}.
\end{itemize}


\begin{enumerate}

\item {\bf Claims}
    \item[] Question: Do the main claims made in the abstract and introduction accurately reflect the paper's contributions and scope?
    \item[] Answer:   \answerYes{}.
    \item[] Justification: The contribution and scope have been reflected in the introduction and abstract, especially in the last part of the introduction.

\item {\bf Limitations}
    \item[] Question: Does the paper discuss the limitations of the work performed by the authors?
    \item[] Answer: \answerYes{} 
    \item[] Justification: The limitation is discussed in the last part of the submission.

\item {\bf Theory assumptions and proofs}
    \item[] Question: For each theoretical result, does the paper provide the full set of assumptions and a complete (and correct) proof?
    \item[] Answer:  \answerNA{}  
    \item[] Justification: Our work does not include theoretical result.

    \item {\bf Experimental result reproducibility}
    \item[] Question: Does the paper fully disclose all the information needed to reproduce the main experimental results of the paper to the extent that it affects the main claims and/or conclusions of the paper (regardless of whether the code and data are provided or not)?
    \item[] Answer: \answerYes{} 
    \item[] Justification: We have disclose all the information for reproduction in the, method, experiment and appendix.

\item {\bf Open access to data and code}
    \item[] Question: Does the paper provide open access to the data and code, with sufficient instructions to faithfully reproduce the main experimental results, as described in supplemental material?
    \item[] Answer: \answerNo{}  
    \item[] Justification: The dataset are open available, and the code will be released after acceptance.

\item {\bf Experimental setting/details}
    \item[] Question: Does the paper specify all the training and test details (e.g., data splits, hyperparameters, how they were chosen, type of optimizer) necessary to understand the results?
    \item[] Answer: \answerYes{} 
    \item[] Justification:  The training and test details have been provided in the experiments and the appendix.

\item {\bf Experiment statistical significance}
    \item[] Question: Does the paper report error bars suitably and correctly defined or other appropriate information about the statistical significance of the experiments?
    \item[] Answer: \answerYes{}  
    \item[] Justification: We have provided statistical significance tests in the appendix.

\item {\bf Experiments compute resources}
    \item[] Question: For each experiment, does the paper provide sufficient information on the computer resources (type of compute workers, memory, time of execution) needed to reproduce the experiments?
    \item[] Answer: \answerYes{} 
    \item[] Justification:  We have provided the compute resources in the experiments.
    
\item {\bf Code of ethics}
    \item[] Question: Does the research conducted in the paper conform, in every respect, with the NeurIPS Code of Ethics \url{https://neurips.cc/public/EthicsGuidelines}?
    \item[] Answer:  \answerYes{} 
    \item[] Justification: This work match with the NeurIPS Code of Ethics in every respect.

\item {\bf Broader impacts}
    \item[] Question: Does the paper discuss both potential positive societal impacts and negative societal impacts of the work performed?
    \item[] Answer: \answerYes{}  
    \item[] Justification:  We have discuss the broader impact in the appendix.
    
\item {\bf Safeguards}
    \item[] Question: Does the paper describe safeguards that have been put in place for responsible release of data or models that have a high risk for misuse (e.g., pre-trained language models, image generators, or scraped datasets)?
    \item[] Answer: \answerNA{} 
    \item[] Justification: This paper poses no such risks.

\item {\bf Licenses for existing assets}
    \item[] Question: Are the creators or original owners of assets (e.g., code, data, models), used in the paper, properly credited and are the license and terms of use explicitly mentioned and properly respected?
    \item[] Answer:  \answerYes{} 
    \item[] Justification:  We have properly credited the original owners when using existing assets.

\item {\bf New assets}
    \item[] Question: Are new assets introduced in the paper well documented and is the documentation provided alongside the assets?
    \item[] Answer:  \answerNA{} 
    \item[] Justification: This paper does not release new assets.

\item {\bf Crowdsourcing and research with human subjects}
    \item[] Question: For crowdsourcing experiments and research with human subjects, does the paper include the full text of instructions given to participants and screenshots, if applicable, as well as details about compensation (if any)? 
    \item[] Answer:  \answerNA{} 
    \item[] Justification: The paper does not involve crowdsourcing nor research with human subjects.

\item {\bf Institutional review board (IRB) approvals or equivalent for research with human subjects}
    \item[] Question: Does the paper describe potential risks incurred by study participants, whether such risks were disclosed to the subjects, and whether Institutional Review Board (IRB) approvals (or an equivalent approval/review based on the requirements of your country or institution) were obtained?
    \item[] Answer:  \answerNA{} 
    \item[] Justification: The paper does not involve crowdsourcing nor research with human subjects.

\item {\bf Declaration of LLM usage}
    \item[] Question: Does the paper describe the usage of LLMs if it is an important, original, or non-standard component of the core methods in this research? Note that if the LLM is used only for writing, editing, or formatting purposes and does \emph{not} impact the core methodology, scientific rigor, or originality of the research, declaration is not required.
    \item[] Answer: \answerYes{}  
    \item[] Justification: We have used the ChatGPT-4o for ``Editing (\emph{e.g.}, grammar, spelling, word choice)'' as stated in OpenReview.

\end{enumerate}